\documentclass[runningheads]{llncs}
\usepackage{graphicx}
\usepackage{comment}
\usepackage{url} 
\usepackage{adjustbox}
\usepackage{caption,subcaption}

\usepackage{array}

\begin{document}
\title{Probing Fairness of Mobile Ocular Biometrics Methods Across Gender on VISOB 2.0 Dataset}
\titlerunning{Fairness of Mobile Ocular Biometrics}
%
\author{Anoop Krishnan$^{*}$ \and
Ali Almadan\thanks{Both the authors contributed equally.} \and
Ajita Rattani\thanks{Corresponding author.}}
\authorrunning{Corresponding author: Ajita Rattani~(ajita.rattani@wichita.edu).}
%
\institute{Department of Electrical Eng. and Computer Science \\
Wichita State University, USA \\
\email{\{axupendrannair,aaalmadan\}@shockers.wichita.edu; ajita.rattani@wichita.edu}\\}

\maketitle             

\begin{abstract}
Recent research has questioned the fairness of face-based recognition and attribute classification methods (such as gender and race) for dark-skinned people and women.
Ocular biometrics in the visible spectrum is an alternate solution over face biometrics, thanks to its accuracy, security, robustness against facial expression, and ease of use in mobile devices. With the recent COVID-19 crisis, ocular biometrics has a further advantage over face biometrics in the presence of a mask. However, fairness of ocular biometrics has not been studied till now. This first study aims to explore the fairness of ocular-based authentication and gender classification methods across males and females. To this aim, VISOB $2.0$ dataset, along with its gender annotations, is used for the fairness analysis of ocular biometrics methods based on ResNet-50, MobileNet-V2 and lightCNN-29 models. Experimental results suggest the equivalent performance of males and females for ocular-based mobile user-authentication in terms of genuine match rate (GMR) at lower false match rates (FMRs) and an overall Area Under Curve (AUC). For instance, an AUC of $0.96$ for females and $0.95$ for males was obtained for lightCNN-29 on an average. However, males significantly outperformed females in deep learning based gender classification models based on ocular-region.
\end{abstract}

\section{Introduction}

With AI and computer vision reaching an inflection point, face biometrics is widely adopted for recognizing identities, surveillance, border control, and mobile user authentication with Apple introducing Face ID moniker in iPhone X\footnote{\url{https://www.apple.com/iphone/}}. The wide-scale integration of biometrics technology in mobile devices facilitate enhanced security in a user login, payment transaction, and eCommerce. 
Over the last few years, \emph{fairness of these automated face-based} recognition~\cite{albiero2020does,9001031,cavazos2019accuracy,singh2020robustness} and gender classification methods have been questioned~\cite{Krishnan2020,Buolamwini18,Muthukumar19} across demographic variations. \emph{Fairness} is defined as the absence of any prejudice or favoritism toward a group based on their inherent or acquired characteristics. Specifically, the majority of these studies raise the concern of higher error rates of face-based recognition and gender\footnote{The term ``sex" would be more appropriate, but in consistency with the existing studies, the term ``gender" is used in this paper.} classification methods\footnote{The term ``methods", ``algorithms" and ``models" are used interchangeably.} for darker-skinned people like African-American, and \textbf{for women}. 

Speculated causes of the difference in the accuracy rates are skin-tone, make-up, facial expression change rate, pose, and illumination variations for face biometrics.
Further, there has been a recent push for alternate solutions for face biometrics due to a significant drop in its performance in the presence of occlusion, such as mask amid COVID-19~\cite{damer2020effect}. Recent $2020$ NIST study~\cite{ngan2020ongoing} suggests the presence of a mask could cause a face recognition system to fail up to $50\%$. 

\begin{figure}[h]
    \centering
    \includegraphics[scale=0.50]{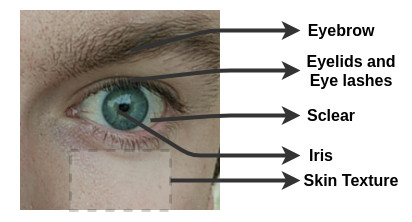}
    \caption{An ocular image labeled with vasculature pattern, eyebrow, eyelids, eyelashes, and periocular skin texture.}
    \label{fig:ocular_region}
\end{figure}

Ocular biometrics in the visible spectrum offers a perfect alternate solution over the face and can be acquired using the front-facing RGB camera already installed in the mobile device~\cite{rattani2019selfie,rattani2016icip,RAJA2020103979}. It comprises of scanning regions in the eye and those around it, i.e., iris, conjunctival and episcleral vasculature and periocular region for person authentication. Figure~\ref{fig:ocular_region} shows an ocular image labeled with vasculature pattern, eyebrow, eyelids, eyelashes, and periocular skin texture. It has obtained significant attention from the research community due to its accuracy, security, robustness against facial expressions, and ease of use in mobile device.  The use of ocular biometrics technology in the mobile device is termed as \emph{mobile ocular biometrics}~\cite{rattani2019selfie,lovisotto2017mobile}. 

With advances in deep learning, deeply coupled autoencoders and convolutional neural networks (CNNs) have been trained from scratch and re-purposed for mobile ocular recognition\footnote{The term ``recognition" and ``user authentication" are used interchangeably.}~\cite{8698586,rattani2019selfie}. Thorough evaluation of fine-tuned CNNs suggests efficacy of ResNet-50, LightCNN and MobileNet in mobile ocular recognition~\cite{8698586}. Datasets such as MICHE-I~\cite{de2015mobile} ($92$ subjects) and VISOB $1.0$~\cite{VISOB} ($550$ subjects) have been assembled for ocular recognition in mobile devices. VISOB $1.0$ dataset was used in the IEEE $2016$ ICIP international competition for mobile ocular biometrics. Studies in~\cite{8440890,9211002,rattani2019selfie} also suggested the efficacy of deep learning-based methods for gender classification from the ocular region in the visible spectrum acquired using a mobile device. The reported results obtained from fine-tuned CNNs suggest that equivalent performance could be obtained in gender classification (with an accuracy of about $85\%$) from the ocular region over face biometrics. 

Recent interest has been in using subject-independent evaluation of these ocular recognition methods where subjects \emph{do not} overlap between the training and testing set to simulate realistic scenarios. To this front, VISOB $2.0$ competition~\cite{VISOB} in IEEE WCCI $2020$ conference has been organized using VISOB $2.0$ database.  VISOB $2.0$~\cite{VISOB} is a new version of the VISOB $1.0$ dataset where the region of interest is extended from the eye (iris, conjunctival, and episcleral vasculature) to periocular (a region encompassing the eye). Further, the evaluation protocol followed is subject-independent, over subject-dependent evaluation in IEEE ICIP VISOB $1.0$ competition~\cite{rattani2016icip}. 
Furthermore, instead of a single frame eye image in VISOB $1.0$~\cite{rattani2016icip}, the data sample consists of a stack of five images captured in burst mode to facilitate multi-frame analysis.

However, to date, \emph{the fairness of these deep learning based mobile ocular biometrics analysis models (such as ResNet-50, LightCNN and MobileNet) has not been evaluated}. It is not known whether ocular biometrics also have an advantage over face biometrics in terms of performance across demographic variations. 
The aim of this paper is to evaluate the fairness of ocular-based recognition and gender classification models across males and females from images acquired using mobile devices. In the context of this study, \emph{fairness} is defined as equivalent error rates (or accuracy rates) for user authentication and gender classification across males and females. \emph{To the best of our knowledge, this is the first study of its kind}. To this aim, the contributions of this paper are as follows:
\begin{itemize}
\item Evaluation of the fairness of deep learning-based methods for mobile ocular-based user authentication across males and females. To this front, performance of the fine-tuned version of ResNet-50~\cite{he2015deep} and lightCNN-29~\cite{wu2018light} have been evaluated using Softmax and cosine loss functions (ArcFace, CosFace, SphereFace, and AdaCos~\cite{almadan2020bwcface}) in three lighting conditions (office, dark and daylight condition). 
\item Evaluation of the fairness of gender classification methods based on ocular region across males and females. The performance of fine-tuned ResNet-50~\cite{he2015deep}, MobileNet-V2~\cite{howard2017mobilenets,sandler2018mobilenetv2} and their ensemble have been evaluated in three lighting conditions across gender.
\end{itemize}

All the experiments are conducted on VISOB $2.0$ dataset~\cite{VISOB}, which facilitates subject-independent evaluation across three lighting conditions; office, daylight, and dark light. 
This paper is organized as follows: section $2$ details deep learning architectures used in this study for ocular analysis. Section $3$ discusses the VISOB $2.0$ training and testing dataset. Sections $4$ and $5$ discuss implementation details, and the obtained results on the fairness of the mobile ocular-based user authentication and gender classification methods, respectively, across males and females.  Conclusions are drawn in section $6$.

\section{Convolutional Neural Networks (CNN) Models Used}
	We used the popular ResNet~\cite{he2015deep}, mobile friendly lightCNN~\cite{wu2018light} and MobileNet~\cite{sandler2018mobilenetv2} based ocular analysis models for our evaluation. Efficacy of these models have already been established for mobile user recognition~\cite{8698586} and gender classification from ocular region~\cite{8440890,9211002,rattani2019selfie}. Experimental results are reported for only two best models for user authentication and gender classification for the sake of space.
	These models (networks) are described below as follows:

	\begin{itemize}
	\item \textbf{ResNet}: ResNet~\cite{he2015deep} is a short form of residual network based on  the  idea  of  ``identity  shortcut  connection'' where input features may skip  certain layers.  
	The residual or shortcut connections introduced in ResNet allow for identity mappings to propagate around multiple nonlinear layers, preconditioning the optimization and alleviating the vanishing gradient problem. In this study, we used ResNet-$50$ model, which has $23.5$M parameters. 
	
	\item \textbf{LightCNN}: This model extensively uses the Max-Feature-Map (MFM) operation instead of ReLu activation, which acts as a feature filter after each convolutional layer~\cite{wu2018light}. The operation takes two feature maps, eliminates the element-wise minimums, and returns element-wise maximums. By doing so across feature channels, only $50\%$ of the information-bearing nodes from each layer reach the next layer. Consequently, during training, each layer is forced to preserve only compact feature maps. Therefore, model parameters and the extracted features are significantly reduced. We used the lightCNN-$29$ model consisting of $12$K parameters in this study.   
	
	\item \textbf{MobileNet}: MobileNet~\cite{howard2017mobilenets,sandler2018mobilenetv2} is one of the most popular mobile-centric deep learning architectures, which is not only small in size but also computationally efficient while achieving high performance. The main idea of MobileNet is that instead of using regular $3\times3$ convolution filters, the operation is split into depth-wise separable $3\times3$ convolution filters followed by $1\times 1$ convolutions. While achieving the same filtering and combination process as a regular convolution, the new architecture requires less number of operations and parameters. In this study, we used MobileNet-V$2$~\cite{sandler2018mobilenetv2} which consist of $3.4$M parameters.
	
	\end{itemize}

\section{VISOB 2.0 Dataset}
In this section, we discuss VISOB $2.0$ dataset along with the experimental protocol.

VISOB $2.0$~\cite{VISOB} is the 2nd version of VISOB $1.0$ dataset used in IEEE WCCI competition $2020$. This publicly available dataset consists of a stack of eye images captured using the burst mode via two mobile devices: Samsung Note 4 and Oppo N1. During the data collection, the volunteers were asked to take their selfie images in two visits, $2$ to $4$ weeks apart from each other. At each visit, the selfie-like images were captured using the front-facing camera of the mobile devices under three lighting conditions (daylight, office light, and dark light) and two sessions (about $10$ to $15$ minutes apart). The stack consisting of five consecutive eye images were extracted from the stack of full-face frames selected such that the correlation coefficient between the center frame and the remaining four images is greater than $90\%$. The face and eye landmarks are detected using the Dlib library~\cite{dlib}. The eye crops were generated such that the width and height of the crop are $2.5\times$ that of eye width.

\vspace{1.5mm} \noindent \textbf{Training and testing subset}: The subset of the VISOB $2.0$ dataset consisting of $150$ subjects each for left and right eye (ocular regions) from two visits are used as the training set. This set was provided to the participants at the IEEE WCCI competition 2020. All the images from visit $1$ and visit $2$ (2-4 weeks apart) under three lighting conditions are included in this training set. 

In order to evaluate the submission for real-life scenarios, $100$ subjects each for left and right eye images are used as the testing set. All the stack of five images per sample from two visits across three lighting conditions is available in this set as well. 
	We used a gender-balanced subset of the dataset for training and testing the models for user-authentication and gender classification (detailed in sections $4$ and $5$). 
	This is in order to mitigate the impact of training and testing set imbalance on the fairness of the models.

	\section{Fairness of Mobile Ocular Recognition Methods Across Gender}
	In this section, we discuss the implementation of the models (networks) for ocular-based user authentication evaluated across males and females. All the implementations are done using Pytorch library (https://pytorch.org/).

\subsection{Network Training and Implementation Details}
ResNet-50~\cite{he2015deep} and lightCNN-29~\cite{wu2018light} are fine-tuned on the training subset of the VISOB $2.0$ dataset~\cite{VISOB} using five different loss functions; Softmax and cosine-based (ArcFace, CosFace, SphereFace, and AdaCos~\cite{almadan2020bwcface}) for the first time for ocular recognition. 
For ResNet-50 based on cosine loss functions (ArcFace, CosFace, SphereFace, and AdaCos~\cite{almadan2020bwcface}), batch normalization, drop-out, and fully connected layers of 2048 and 512 are added after the last convolutional layer. This is followed by the final output layer. In case of lightCNN-29, the layers added after the last pooling layer: batch normalization, drop-out, and a fully connected layer of 128 x 128 x 8 and 512, and followed by the output layer.
 The angular margins are set to $0.50$, $0.40$, and $4.0$ for ArcFace, CosFace, and SphereFace.  AdaCos adjusts its scale parameter automatically. SphereFace obtained equivalent performance with AdaCos and results are not included due to space constraints. The ResNet-50 network is trained using Adam optimizer with a batch size of $128$ for $15$ iterations, and lightCNN-29 using stochastic gradient decent optimizer with a batch size of $64$ and the same number of iterations as ResNet-50. The learning rate was set to $0.001$. 

We used a gender-balanced subset of the training set of these models. Following IEEE WCCI competition protocol, left and right eye images are treated as different identities. To this aim, $288$ subjects consisting of left and right eye individually (with $50\%$ male and $50\%$ female distribution) are randomly chosen. The training set consists of $64$K ocular images from all the lighting conditions and the two visits. In particular, the number of samples from visit 1 is around $32$K and $40$K from visit 2 where each subject has $500$ images. The models are trained and validated on a split of $80/20$ using samples from both the visits and across all the lighting conditions for left and right eyes\footnote{The term ``eye" and ``ocular region" are used interchangeably.} are used for training the models. 
	
The trained models are evaluated using a subject-independent testing set of VISOB $2.0$. For the purpose of this study, we used a gender-balanced version of the test set as well.
This results in a total of $21$K ocular images for each gender for Oppo device, and $15$K images for Note-4 from $86$ subjects for all the three lighting conditions. 
The deep features of size $512-D$ are extracted from the fully connected layers of these trained models for the evaluation. The deep features from the samples in visit 1 and visit 2 are chosen as the template and query pairs, respectively. The scores are computed in a pairwise fashion over a stack of images and are averaged per template-query pair.  Cosine similarity is used to compute scores between deep features from a pair of template-query pair.

\subsection{Experimental Results}
In this section, we compare the verification performance of the ResNet-50 and lightCNN-29 models trained using multiple loss functions on the gender-balanced subset. Both models are evaluated in the same lighting conditions for both left and right eye regions. Table ~\ref{light-results}  and ~\ref{res-results} compare the Equal Error Rate (EER) and Genuine Match Rate (GMR) across males and females at $1^{-4}$, $1^{-3}$, and $1^{-2}$ FMRs. 

\par Tables ~\ref{light-results} and ~\ref{res-results} shows Equal Error Rates (EER) and Genuine Match Rates (GMR) at three different False Match Rates (FMR). In general, both models performed better for females than males. For instance, lightCNN-29 obtained an average EER of $9.94$ for females and $11.16$ for males, respectively. ResNet-50 obtained higher EER of $17.47$ for females and $20.15$ for males over ResNet-50. However, both the genders obtained similar Genuine Match Rate (GMR) averaged over both the models. For instance, lightCNN-29 obtained $36.94$ and $37.69$ for females and males at GMR@$1^{-4}$FMR, respectively. At the same FMR, ResNet-50 obtained GMR of $6.43$ and about $5.44$ for females and males, respectively. However, females tends to outperform males remarkably at GMR@$1^{-2}$FMR. This can be noticed for lightCNN-29, which obtained an average of $71.00$ GMR on females compared to $66.87$ on males. Similarly, a difference of about $7\%$ was obtained for ResNet-50 across females and males.

\par Across lighting conditions, females obtained equivalent EERs across dark and office lighting conditions. For lightCNN model, females obtained an average EER of $9.94$ for dark and $9.43$ for office light, respectively. For ResNet-50, EER of $16.04$ and $16.54$ for dark and office light, respectively. On the other hand, males performed the best in dark conditions for both models compared to other lighting conditions. This can be observed as the EERs increased by about $4.5$\% for daylight and $3$\% for office light compared to dark lighting condition.

\par The performance across gender for different loss functions varied depending on the lighting conditions and the CNN model. AdaCos loss function results in lower EER for females than males for both the models and across all the lighting conditions. For instance, in dark condition, an overall increase in EER of $0.964$ for lightCNN-29 and $2.27$ for ResNet-50 was observed. ArcFace also performed better for females than males except in the case of lightCNN-29 in dark conditions where the EER of females increased by $0.88$. For other lighting conditions, males obtained an average increase of $2.67$  in EER over females. At GMR@$1^{-2}$FMR, females obtained higher performance across many loss functions and lighting conditions, yet the performance for females dropped uniquely across all loss functions in the dark conditions for lightCNN-29, except for AdaCos. The average drop at GMR@$1^{-2}$FMR was about $2.79 \%$ for ArcFace, CosFace, SphereFace, and Softmax where AdaCos declined for males by only 1.14\%. In daylight and office light conditions, a constant increase in GMR@$1^{-2}$FMR for females was noticed. 

\par Generally, the average AUC of both genders were equivalent for lightCNN-29 (\textbf{0.96} for females and \textbf{0.95} for males). In the case of ResNet-50, the average AUC for females was \textbf{0.90}, whereas males obtained an AUC of \textbf{0.87}.


\begin{table} [ht]
\centering
\caption{\small{EER(\%), GMR@$1^{-4}$FMR, GMR@$1^{-3}$FMR, and GMR@$1^{-2}$FMR for lightCNN-29 model (trained on gender balanced subset of VISOB 2.0) for five loss functions and evaluated in different light conditions for males (M) and females (F) for mobile user authentication. Gender-balanced training and testing subset of VISOB 2.0 are used.} }
\resizebox{\textwidth}{!}{
\begin{tabular}{|p{1.5cm}|p{1.5cm}|p{1.5cm}|p{1.5cm}|p{1.5cm}|p{1.5cm}|p{1.5cm}|p{1.5cm}|p{1.5cm}|p{1.5cm}|p{1.5cm}|} \hline
{\textbf{Loss Function}} & {\textbf{Light Condition}} & \textbf{Eye} & \multicolumn{2}{c|}{\textbf{EER(\%)~~}} & \multicolumn{2}{c|}{\textbf{GMR(/\%)@1-4FMR}} & \multicolumn{2}{c|}{\textbf{GMR(\%)@1-3FMR}} & \multicolumn{2}{c|}{\textbf{GMR(\%)@1-2FMR}}  \\ 
\cline{4-11}
                                        &                                           &                      & \multicolumn{8}{c|}{\textbf{lightCNN-29 - Note-4 }}                                                                                                                                     \\ 
\cline{4-11}
                                        &                                           &                      & \textbf{M}     & \textbf{F}                                & \textbf{M}     & \textbf{F}                                      & \textbf{M}     & \textbf{F}                                    & \textbf{M}     & \textbf{F}                                     \\ 
\hline
{\textbf{AdaCos}}        & {\textbf{Office}}          & L                    & 7.73  & 3.43                            & 29.04 & 50.33                                 & 46.25 & 58.89                                & 72.06 & 87.02                                 \\ 
\cline{3-11}
                                        &                                           & R                    & 4.26  & 4.15                            & 67.25 & 56.76                                 & 76.56 & 69.48                                & 87.53 & 87.15                                 \\ 
\cline{2-11}
                                        & {\textbf{Daylight}}        & L                    & 4.93  & 3.19                            & 55.96 & 65.42                                 & 71.90 & 78.39                                & 84.29 & 92.42                                 \\ 
\cline{3-11}
                                        &                                           & R                    & 4.66  & 2.84                            & 76.40 & 64.79                                 & 81.69 & 80.12                                & 90.37 & 93.84                                 \\ 
\cline{2-11}
                                        & {\textbf{Dark}}            & L                    & 6.39  & 2.20                            & 38.99 & 53.36                                 & 56.43 & 65.61                                & 78.93 & 93.15                                 \\ 
\cline{3-11}
                                        &                                           & R                    & 5.54  & 2.61                            & 72.22 & 79.85                                 & 77.82 & 86.75                                & 85.79 & 95.38                                 \\ 
\hline
{\textbf{ArcFace}}       & {\textbf{Office}}          & L                    & 17.21 & 14.25                           & 11.84 & 12.85                                 & 25.65 & 28.52                                & 49.17 & 44.60                                 \\ 
\cline{3-11}
                                        &                                           & R                    & 13.12 & 11.04                           & 39.15 & 11.71                                 & 48.30 & 29.85                                & 64.64 & 54.08                                 \\ 
\cline{2-11}
                                        & {\textbf{Daylight}}        & L                    & 9.01  & 5.34                            & 39.38 & 31.49                                 & 59.12 & 39.00                                & 78.93 & 82.15                                 \\ 
\cline{3-11}
                                        &                                           & R                    & 4.53  & 6.16                            & 64.48 & 55.82                                 & 71.27 & 68.27                                & 90.21 & 79.52                                 \\ 
\cline{2-11}
                                        & {\textbf{Dark}}            & L                    & 11.10 & 7.87                            & 17.25 & 21.84                                 & 38.26 & 40.77                                & 63.39 & 76.76                                 \\ 
\cline{3-11}
                                        &                                           & R                    & 6.37  & 9.63                            & 53.03 & 39.94                                 & 60.28 & 65.16                                & 77.98 & 77.05                                 \\ 
\hline
{\textbf{CosFace}}       & {\textbf{Office}}          & L                    & 10.37 & 11.53                           & 49.17 & 34.02                                 & 61.38 & 46.30                                & 74.95 & 68.42                                 \\ 
\cline{3-11}
                                        &                                           & R                    & 9.36  & 7.93                            & 60.18 & 54.58                                 & 70.92 & 73.75                                & 82.39 & 82.53                                 \\ 
\cline{2-11}
                                        & {\textbf{Daylight}}        & L                    & 11.38 & 8.81                            & 51.65 & 34.30                                 & 58.26 & 40.49                                & 72.75 & 74.79                                 \\ 
\cline{3-11}
                                        &                                           & R                    & 6.16  & 8.22                            & 60.37 & 58.73                                 & 71.10 & 65.82                                & 86.97 & 85.46                                 \\ 
\cline{2-11}
                                        & {\textbf{Dark}}            & L                    & 22.93 & 19.96                           & 8.35  & 17.53                                 & 14.40 & 25.40                                & 30.09 & 42.92                                 \\ 
\cline{3-11}
                                        &                                           & R                    & 17.71 & 16.45                           & 25.96 & 10.01                                 & 33.49 & 25.21                                & 51.93 & 48.91                                 \\ 
\hline
{\textbf{Softmax}}       & {\textbf{Office}}          & L                    & 11.93 & 9.56                            & 31.56 & 13.03                                 & 44.13 & 34.68                                & 64.31 & 68.98                                 \\ 
\cline{3-11}
                                        &                                           & R                    & 11.38 & 10.57                           & 45.87 & 39.66                                 & 60.64 & 57.88                                & 74.50 & 76.96                                 \\ 
\cline{2-11}
                                        & {\textbf{Daylight}}        & L                    & 5.95  & 9.27                            & 47.19 & 30.96                                 & 62.45 & 44.35                                & 78.35 & 75.54                                 \\ 
\cline{3-11}
                                        &                                           & R                    & 6.49  & 3.92                            & 64.61 & 64.26                                 & 71.54 & 71.00                                & 90.37 & 87.38                                 \\ 
\cline{2-11}
                                        & {\textbf{Dark}}            & L                    & 3.14  & 5.02                            & 72.29 & 45.10                                 & 81.71 & 60.51                                & 91.77 & 82.27                                 \\ 
\cline{3-11}
                                        &                                           & R                    & 4.44  & 2.60                            & 73.16 & 78.29                                 & 81.28 & 84.56                                & 90.58 & 93.42                                 \\

\cline{2-11}
                                        & {\textbf{Dark}}            & L                    & 8.69  & 8.01                            & 34.96 & 33.36                                 & 43.51 & 54.97                                & 77.06 & 77.56                                 \\ 
\cline{3-11}
                                        &                                           & R                    & 3.35  & 6.90                            & 73.16 & 61.76                                 & 79.11 & 70.53                                & 90.37 & 81.66                                 \\ 
\hline
                                        &                                           &                      & \multicolumn{8}{c|}{\textbf{lightCNN-29 - Oppo}}                                                                                                                                       \\ 
\hline
{\textbf{AdaCos}}        & {\textbf{Office}}          & L                    & 9.68  & 10.46                           & 28.31 & 27.10                                 & 36.20 & 47.07                                & 63.16 & 71.48                                 \\ 
\cline{3-11}
                                        &                                           & R                    & 14.32 & 11.08                           & 17.99 & 28.10                                 & 27.04 & 43.57                                & 51.96 & 70.02                                 \\ 
\cline{2-11}
                                        & {\textbf{Daylight}}        & L                    & 12.42 & 10.71                           & 23.82 & 21.04                                 & 41.44 & 42.17                                & 60.02 & 69.51                                 \\ 
\cline{3-11}
                                        &                                           & R                    & 15.51 & 11.87                           & 21.20 & 35.32                                 & 30.35 & 43.68                                & 46.90 & 63.21                                 \\ 
\cline{2-11}
                                        & {\textbf{Dark}}            & L                    & 16.10 & 10.98                           & 28.58 & 20.04                                 & 38.90 & 39.62                                & 52.30 & 70.83                                 \\ 
\cline{3-11}
                                        &                                           & R                    & 13.39 & 12.22                           & 25.80 & 22.41                                 & 35.37 & 36.65                                & 60.39 & 60.71                                 \\ 
\hline
{\textbf{ArcFace}}       & {\textbf{Office}}          & L                    & 21.24 & 18.72                           & 12.53 & 12.12                                 & 18.91 & 26.17                                & 30.09 & 43.13                                 \\ 
\cline{3-11}
                                        &                                           & R                    & 20.42 & 19.70                           & 9.36  & 10.22                                 & 17.68 & 22.25                                & 32.01 & 44.22                                 \\ 
\cline{2-11}
                                        & {\textbf{Daylight}}        & L                    & 16.42 & 14.05                           & 19.83 & 18.14                                 & 29.12 & 31.44                                & 49.43 & 54.34                                 \\ 
\cline{3-11}
                                        &                                           & R                    & 16.71 & 13.57                           & 11.74 & 17.09                                 & 19.23 & 28.63                                & 32.57 & 49.23                                 \\ 
\cline{2-11}
                                        & {\textbf{Dark}}            & L                    & 11.93 & 7.97                            & 21.36 & 32.26                                 & 37.21 & 56.92                                & 59.98 & 78.64                                 \\ 
\cline{3-11}
                                        &                                           & R                    & 12.03 & 8.48                            & 39.81 & 33.17                                 & 49.35 & 57.27                                & 65.31 & 79.59                                 \\ 
\hline
{\textbf{CosFace}}       & {\textbf{Office}}          & L                    & 13.29 & 9.39                            & 22.23 & 55.53                                 & 39.81 & 66.19                                & 61.53 & 79.11                                 \\ 
\cline{3-11}
                                        &                                           & R                    & 12.25 & 7.67                            & 49.53 & 40.47                                 & 62.43 & 50.93                                & 71.69 & 74.08                                 \\ 
\cline{2-11}
                                        & {\textbf{Daylight}}        & L                    & 14.12 & 10.55                           & 30.37 & 45.86                                 & 41.93 & 58.66                                & 60.99 & 77.61                                 \\ 
\cline{3-11}
                                        &                                           & R                    & 11.64 & 10.99                           & 53.57 & 44.15                                 & 60.95 & 55.69                                & 71.76 & 70.83                                 \\ 
\cline{2-11}
                                        & {\textbf{Dark}}            & L                    & 20.21 & 15.50                           & 15.09 & 8.28                                  & 19.09 & 26.20                                & 34.40 & 53.23                                 \\ 
\cline{3-11}
                                        &                                           & R                    & 15.22 & 13.08                           & 19.20 & 22.24                                 & 26.59 & 32.46                                & 47.98 & 54.82                                 \\ 
\hline
{\textbf{Softmax}}       & {\textbf{Office}}          & L                    & 18.14 & 15.58                           & 27.70 & 47.25                                 & 39.63 & 58.34                                & 52.27 & 71.30                                 \\ 
\cline{3-11}
                                        &                                           & R                    & 12.90 & 9.31                            & 41.43 & 43.56                                 & 50.86 & 53.47                                & 64.48 & 70.31                                 \\ 
\cline{2-11}
                                        & {\textbf{Daylight}}        & L                    & 9.93  & 8.56                            & 34.41 & 29.67                                 & 47.85 & 41.97                                & 73.88 & 70.33                                 \\ 
\cline{3-11}
                                        &                                           & R                    & 10.48 & 7.24                            & 32.50 & 42.84                                 & 46.94 & 52.25                                & 61.07 & 74.98                                 \\ 
\cline{2-11}
                                        & {\textbf{Dark}}            & L                    & 6.38  & 8.84                            & 20.69 & 44.46                                 & 42.74 & 60.23                                & 77.51 & 74.45                                 \\ 
\cline{3-11}
                                        &                                           & R                    & 8.13  & 9.17                            & 40.44 & 39.50                                 & 53.73 & 57.20                                & 77.10 & 76.55                                 \\ 
\hline

\end{tabular}}
 
\label{light-results}
\end{table}

\begin{table} [ht]
\centering
\caption{\small {EER(\%), GMR@$1^{-4}$FMR, GMR@$1^{-3}$FMR, and GMR@$1^{-2}$FMR for ResNet-50 model using five loss functions and evaluated in different light conditions across males (M) and females (F) for mobile user authentication. Gender-balanced training and testing subset of VISOB 2.0 are used.} } 
\resizebox{\textwidth}{!}{
\begin{tabular}{|p{1.5cm}|p{1.5cm}|p{1.5cm}|p{1.5cm}|p{1.5cm}|p{1.5cm}|p{1.5cm}|p{1.5cm}|p{1.5cm}|p{1.5cm}|p{1.5cm}|} \hline
{\textbf{Loss Function}} & {\textbf{Light Condition}} & \textbf{Eye} & \multicolumn{2}{c|}{\textbf{EER(\%)~~}} & \multicolumn{2}{c|}{\textbf{GMR(/\%)@1-4FMR}} & \multicolumn{2}{c|}{\textbf{GMR(\%)@1-3FMR}} & \multicolumn{2}{c|}{\textbf{GMR(\%)@1-2FMR}}  \\ 

\hline

                                        &                                           &                               & \multicolumn{8}{c|}{\textbf{ResNet-50 - Note-4}}                                                                                                                                          \\ \cline{4-11} 
                                        &                                           &                               & \textbf{M}                 & \textbf{F}                 & \textbf{M}                     & \textbf{F}                    & M                     & \textbf{F}                     & \textbf{M}                        & \textbf{F}                       \\ \hline
{\textbf{AdaCos}}        & {\textbf{Office}}          & L                             & 19.65             & 12.38             & 0.00                  & 4.48                 & 4.74                  & 12.38                & 27.86                    & 41.63                   \\ \cline{3-11} 
                                        &                                           & R                             & 10.45             & 8.45              & 9.87                  & 1.00                 & 28.57                 & 15.66                & 58.41                    & 59.84                   \\ \cline{2-11} 
                                        & {\textbf{Daylight}}        & L                             & 24.80             & 16.24             & 0.28                  & 16.49                & 3.67                  & 24.27                & 19.08                    & 48.55                   \\ \cline{3-11} 
                                        &                                           & R                             & 19.36             & 13.69             & 5.41                  & 0.00                 & 18.35                 & 2.27                 & 41.93                    & 31.35                   \\ \cline{2-11} 
                                        & {\textbf{Dark}}            & L                             & 17.54             & 13.61             & 4.65                  & 5.24                 & 7.36                  & 14.88                & 19.05                    & 40.39                   \\ \cline{3-11} 
                                        &                                           & R                             & 18.29             & 14.65             & 5.19                  & 8.54                 & 17.97                 & 17.87                & 51.62                    & 46.87                   \\ \hline
{\textbf{ArcFace}}       & {\textbf{Office}}          & L                             & 25.36             & 18.71             & 1.10                  & 2.24                 & 5.21                  & 7.18                 & 22.65                    & 25.76                   \\ \cline{3-11} 
                                        &                                           & R                             & 19.02             & 11.24             & 12.23                 & 3.15                 & 18.94                 & 26.97                & 38.12                    & 61.18                   \\ \cline{2-11} 
                                        & {\textbf{Daylight}}        & L                             & 21.83             & 15.87             & 1.10                  & 6.00                 & 6.33                  & 13.50                & 34.59                    & 45.36                   \\ \cline{3-11} 
                                        &                                           & R                             & 18.94             & 16.52             & 0.18                  & 6.70                 & 7.16                  & 27.10                & 30.83                    & 50.90                   \\ \cline{2-11} 
                                        & {\textbf{Dark}}            & L                             & 21.64             & 17.35             & 5.41                  & 4.64                 & 10.17                 & 11.52                & 29.55                    & 30.59                   \\ \cline{3-11} 
                                        &                                           & R                             & 15.27             & 11.99             & 4.55                  & 5.02                 & 20.89                 & 19.20                & 46.43                    & 48.12                   \\ \hline
{\textbf{CosFace}}       & {\textbf{Office}}          & L                             & 19.42             & 11.73             & 2.68                  & 8.83                 & 5.84                  & 15.68                & 19.10                    & 49.47                   \\ \cline{3-11} 
                                        &                                           & R                             & 12.31             & 12.38             & 0.95                  & 20.62                & 14.44                 & 31.79                & 55.56                    & 56.43                   \\ \cline{2-11} 
                                        & {\textbf{Daylight}}        & L                             & 24.60             & 17.06             & 3.76                  & 10.31                & 9.27                  & 18.93                & 26.61                    & 44.89                   \\ \cline{3-11} 
                                        &                                           & R                             & 21.96             & 19.64             & 1.19                  & 3.78                 & 2.75                  & 9.92                 & 26.97                    & 38.05                   \\ \cline{2-11} 
                                        & {\textbf{Dark}}            & L                             & 15.58             & 19.45             & 0.65                  & 0.90                 & 4.44                  & 3.89                 & 27.38                    & 19.00                   \\ \cline{3-11} 
                                        &                                           & R                             & 15.19             & 16.22             & 10.82                 & 6.97                 & 18.40                 & 21.00                & 48.70                    & 41.85                   \\ \hline
{\textbf{Softmax}}       & {\textbf{Office}}          & L                             & 20.44             & 7.97              & 8.13                  & 13.11                & 20.36                 & 32.54                & 43.96                    & 62.12                   \\ \cline{3-11} 
                                        &                                           & R                             & 10.90             & 9.24              & 15.63                 & 7.70                 & 31.49                 & 20.88                & 59.19                    & 60.84                   \\ \cline{2-11} 
                                        & {\textbf{Daylight}}        & L                             & 15.59             & 18.27             & 11.28                 & 21.37                & 18.90                 & 29.80                & 45.50                    & 47.80                   \\ \cline{3-11} 
                                        &                                           & R                             & 12.66             & 13.31             & 33.21                 & 26.63                & 38.81                 & 38.81                & 55.87                    & 65.72                   \\ \cline{2-11} 
                                        & {\textbf{Dark}}            & L                             & 16.12             & 10.11             & 19.16                 & 4.94                 & 28.68                 & 19.37                & 44.70                    & 52.88                   \\ \cline{3-11} 
                                        &                                           & R                             & 8.66              & 10.97             & 20.02                 & 20.92                & 33.87                 & 32.84                & 65.48                    & 58.86                   \\ \hline
                                        &                                           &                               & \multicolumn{8}{c|}{\textbf{ResNet-50 - Oppo}}                                                                                                                                           \\ \hline
{\textbf{AdaCos}}        & {\textbf{Office}}          & L                             & 22.00             & 20.65             & 1.46                  & 8.31                 & 11.78                 & 18.39                & 32.52                    & 36.09                   \\ \cline{3-11} 
                                        &                                           & R                             & 27.25             & 19.59             & 6.10                  & 3.00                 & 10.91                 & 14.86                & 23.32                    & 38.69                   \\ \cline{2-11} 
                                        & {\textbf{Daylight}}        & L                             & 28.76             & 22.31             & 0.68                  & 3.25                 & 1.66                  & 12.05                & 19.31                    & 28.97                   \\ \cline{3-11} 
                                        &                                           & R                             & 23.05             & 25.29             & 1.62                  & 2.73                 & 10.59                 & 12.45                & 30.08                    & 35.16                   \\ \cline{2-11} 
                                        & {\textbf{Dark}}            & L                             & 18.43             & 17.48             & 4.11                  & 3.87                 & 14.19                 & 9.82                 & 43.14                    & 42.42                   \\ \cline{3-11} 
                                        &                                           & R                             & 17.70             & 17.14             & 0.56                  & 3.42                 & 13.73                 & 5.11                 & 32.62                    & 32.18                   \\ \hline
{\textbf{ArcFace}}       & {\textbf{Office}}          & L                             & 24.21             & 23.91             & 7.29                  & 4.20                 & 19.61                 & 15.42                & 36.36                    & 33.72                   \\ \cline{3-11} 
                                        &                                           & R                             & 25.24             & 20.93             & 13.19                 & 3.11                 & 18.56                 & 12.66                & 30.92                    & 36.41                   \\ \cline{2-11} 
                                        & {\textbf{Daylight}}        & L                             & 27.61             & 28.74             & 1.19                  & 1.39                 & 7.89                  & 8.20                 & 21.90                    & 26.28                   \\ \cline{3-11} 
                                        &                                           & R                             & 23.27             & 24.49             & 3.46                  & 4.40                 & 10.66                 & 16.45                & 32.13                    & 34.68                   \\ \cline{2-11} 
                                        & {\textbf{Dark}}            & L                             & 22.50             & 18.66             & 6.10                  & 3.83                 & 13.32                 & 11.49                & 27.31                    & 31.50                   \\ \cline{3-11} 
                                        &                                           & R                             & 21.39             & 16.46             & 7.86                  & 2.41                 & 14.09                 & 9.01                 & 29.64                    & 37.05                   \\ \hline
{\textbf{CosFace}}       & {\textbf{Office}}          & L                             & 22.99             & 24.08             & 0.81                  & 0.02                 & 1.24                  & 0.70                 & 29.07                    & 16.79                   \\ \cline{3-11} 
                                        &                                           & R                             & 24.21             & 22.34             & 7.91                  & 0.72                 & 12.51                 & 2.42                 & 26.99                    & 34.56                   \\ \cline{2-11} 
                                        & {\textbf{Daylight}}        & L                             & 31.41             & 23.50             & 2.09                  & 2.54                 & 9.55                  & 9.75                 & 24.03                    & 28.38                   \\ \cline{3-11} 
                                        &                                           & R                             & 25.50             & 26.40             & 2.92                  & 1.55                 & 15.13                 & 4.44                 & 31.81                    & 32.58                   \\ \cline{2-11} 
                                        & {\textbf{Dark}}            & L                             & 19.62             & 20.92             & 5.98                  & 3.71                 & 15.43                 & 23.96                & 36.04                    & 43.56                   \\ \cline{3-11} 
                                        &                                           & R                             & 22.40             & 20.55             & 0.36                  & 2.37                 & 7.38                  & 14.92                & 27.82                    & 35.96                   \\ \hline
{\textbf{Softmax}}       & {\textbf{Office}}          & L                             & 21.40             & 17.24             & 8.81                  & 9.01                 & 17.61                 & 18.81                & 38.74                    & 41.59                   \\ \cline{3-11} 
                                        &                                           & R                             & 23.69             & 17.95             & 6.98                  & 9.20                 & 10.96                 & 20.09                & 25.59                    & 44.10                   \\ \cline{2-11} 
                                        & {\textbf{Daylight}}        & L                             & 20.10             & 16.01             & 8.29                  & 13.48                & 21.65                 & 26.91                & 40.85                    & 48.55                   \\ \cline{3-11} 
                                        &                                           & R                             & 20.82             & 16.80             & 7.78                  & 8.13                 & 18.88                 & 18.31                & 42.76                    & 42.73                   \\ \cline{2-11} 
                                        & {\textbf{Dark}}            & L                             & 11.56             & 12.80             & 13.04                 & 14.87                & 26.36                 & 27.55                & 49.72                    & 50.00                   \\ \cline{3-11} 
                                        &                                           & R                             & 17.30             & 10.92             & 10.71                 & 18.10                & 22.10                 & 33.39                & 43.49                    & 53.70                   \\ \hline

\end{tabular}}
\label{res-results}
\end{table}

\section{Fairness of Mobile Ocular-based Gender Classification Methods}
In this section, we evaluate the fairness of the gender classification models based on the ocular region. Following the studies in~\cite{8440890,9211002,rattani2019selfie}, we fine-tuned ResNet-$50$, MobileNet-v$2$ and their ensemble for gender classification. Next, we discuss the implementation details and the obtained results. 

\begin{table*}[]
\centering
    \caption{Almost gender-balanced subset of VISOB $2.0$ dataset subset used for training gender classification models.}
\begin{adjustbox}{width=0.5\columnwidth}
\begin{tabular}{|c|c|c|c|c|c|c|}
\hline
\multicolumn{1}{|c|}{{\textbf{\begin{tabular}[c]{@{}c@{}}Lighting \\ Condition\end{tabular}}}} & \multicolumn{2}{c|}{\textbf{Left Eye}} & \multicolumn{2}{c|}{\textbf{Right Eye}} \\ \cline{2-5} 
\multicolumn{1}{|c|}{}                                                                                       & \textbf{M}       & \textbf{F}      & \textbf{M}       & \textbf{F}       \\ \hline
Dark                                                                                               & 35,917            & 31,023           & 35,917            & 31,023            \\ \hline
Daylight                                                                                            & 36,095            & 30,742           & 36,095            & 30,742            \\ \hline
Office                                                                                              & 44,669            & 38,424           & 44,669            & 38,424            \\ \hline
\end{tabular}%
\end{adjustbox}
\label{Train_set}
\end{table*}   

\begin{table*}[]
\centering
    \caption{Gender-balanced subject independent testing subset of VISOB $2.0$ dataset used for gender classification model evaluation.}
\begin{adjustbox}{width=0.75\columnwidth}

\begin{tabular}{|l|c|c|c|c|l|l|l|l|}
\hline
                                                                                            & \multicolumn{4}{c|}{\textbf{NOTE4}}                                      & \multicolumn{4}{c|}{\textbf{Oppo}}                                                                                                    \\ \hline
\multicolumn{1}{|c|}{\textbf{\begin{tabular}[c]{@{}c@{}}Lighting\\ Condition\end{tabular}}} & \multicolumn{2}{c|}{\textbf{Left Eye}} & \multicolumn{2}{c|}{\textbf{Right Eye}} & \multicolumn{2}{c|}{\textbf{Left Eye}}                                & \multicolumn{2}{c|}{\textbf{Right Eye}}                               \\ \hline
\textit{}                                                                                   & \textbf{M}       & \textbf{F}      & \textbf{M}       & \textbf{F}       & \multicolumn{1}{c|}{\textbf{M}} & \multicolumn{1}{c|}{\textbf{F}} & \multicolumn{1}{c|}{\textbf{M}} & \multicolumn{1}{c|}{\textbf{F}} \\ \hline
Dark                                                                               & 1019             & 1020            & 1020             & 1020             & 1525                            & 1525                            & 1525                            & 1525                            \\ \hline
Daylight                                                                           & 1300             & 1300            & 1300             & 1300             & 1590                            & 1590                            & 1590                            & 1590                            \\ \hline
Office                                                                             & 1485             & 1485            & 1485             & 1485             & 2515                            & 2515                            & 2515                            & 2515                            \\ \hline
\end{tabular}%
\end{adjustbox}
\label{Test_set}
\end{table*}

\subsection{Network training and Implementation details}

\par ResNet-$50$ and MobileNet-V$2$ CNN models are fine-tuned on training subset of VISOB $2.0$ dataset. We also evaluated ensemble of ResNet-$50$ and MobileNet-V$2$ models.
 For fine-tuning ResNet-$50$ and MobileNet-V$2$, fully connected layers of $512$ and $512$ were added after the last convolutional layer, followed by the final output layer. Ensemble of ResNet-$50$ and MobileNet-V$2$ was obtained by concatenating their first fully connected layers (of size $1024$), followed by the final output layer.  The above models were trained using an Adamax optimizer\footnote{\url{https://pytorch.org/docs/stable/optim.html}} on a batch size of $128$ for $100$ epochs using an early stopping mechanism on the validation set ($80$-$20$ split of training and validation).  The learning rate was set equal to 1e-4 and decay of 5e-4. In order to mitigate the impact of imbalanced training and evaluation set on the fairness of the models. We used an almost gender balanced subset of the VISOB $2.0$ training set (shown in Table~\ref{Train_set}) for these models training. Samples across both the visits (1 and 2) and all three lighting conditions are used all together to train the models for left and right eye, individually. Validation accuracy of about $90\%$ was obtained for most of the cases. The trained models are evaluated on a subject independent gender-balanced testing subset of the VISOB $2.0$ dataset shown in Table~\ref{Test_set}. Results are reported in terms of accuracy values across gender and lighting conditions for the left and right eye, individually. Further, false positive rate (FPR), indicating females misclassified as males, and false negative rate (FNR), indicating males misclassified as females, are also reported for further insight.
    
\subsection{Experimental Results}
\par In this section, we report the gender classification accuracy of the ocular-based models across males and females.
    
\par Tables~\ref{left_Note4},~\ref{left_Oppo},~\ref{Right_Note4},~\ref{Right_Oppo} shows the accuracy of the fine-tuned ResNet-$50$, MobileNet-V$2$, and their ensemble across gender and lighting conditions for left and right ocular images acquired using Note-4 and Oppo, individually. FPR and FNR are also reported in these tables.
 For Note-4, the average gender classification accuracy across lighting conditions is $79.72\%$, $81.62\%$, and $84.02\%$ for ResNet-50, MobileNet-V2 and their ensemble, respectively, when trained and tested on the left ocular region (as can be seen from Table~\ref{left_Note4}). Similarly, for the right ocular region, Resnet-$50$ has the highest average accuracy of $83.98\%$, followed by ensemble with an accuracy of $82.46$\% and MobileNet-v$2$ with an average accuracy of $81.02$\% (as can be seen from Table~\ref{Right_Note4}).

\par Across gender for left ocular region acquired using Note4; males obtained the highest average accuracy of $94.07\%$ and the lowest of $89.4\%$ for ResNet-$50$ and MobileNet-V$2$, respectively. However, females obtained the highest of $75.03$\% and the lowest of $64.2$\% for Ensemble and ResNet-$50$, respectively, averaged over three different lighting conditions (Table~\ref{left_Note4}). Similarly, for the right ocular region, males obtained the highest average accuracy of $94.9$\% and the lowest of $89.27\%$ for ResNet-$50$ and Ensemble, respectively. However, females obtained the  highest of $77.93$\% and the lowest of $65.83$\% for MobileNet-v$2$ and ResNet-$50$, respectively (see Table~\ref{Right_Note4}).

\par Average difference in the accuracy between males and females is $21.21\%$ for left ocular images acquired using Note-4.  The average difference in the accuracy between males and females is $21.75\%$ for right ocular images acquired using Note-4.  

\par For Oppo, the average gender classification across different lighting conditions is $79.17$\%, $77.8$\%, and $80.42$\% for ResNet-$50$, MobileNet-V$2$ and their ensemble, respectively, when trained and tested on left ocular region (as can be seen from Table~\ref{left_Oppo}). Similarly for right ocular region, ResNet-$50$, MobileNet-v$2$ and their ensemble obtained $80.49$\%, $84.89$\% and $83.09$\%, respectively (see Table~\ref{Right_Oppo}).

\par Across gender for left ocular region acquired using Oppo; males obtained the highest average accuracy of $95.4\%$ and the lowest of $94.54\%$ for  MobileNet-V$2$ and ResNet-$50$, respectively. However, females obtained the highest of $65.13$\% and the lowest of $59.19$\% for Ensemble and MobileNet-V$2$, respectively, averaged over three different lighting conditions (refer Table~\ref{left_Oppo}). Similarly, for the right ocular region, males obtained the highest average accuracy of $93.19$\% and the lowest of $86.45\%$ for ResNet-$50$ and MobileNet-v$2$, respectively. However, females obtained the  highest of $86.38$\% and the lowest of $67.23$\% for MobileNet-v$2$ and ResNet-$50$, respectively (see Table~\ref{Right_Oppo}). Better classification accuracy for Oppo device is due to the higher resolution images of better quality compared to Note-4. Also, in general, higher accuracy rates are obtained for samples acquired in controlled lighting conditions, i.e., office light.

\par Average difference in the accuracy between males and females is $32.7\%$ for left ocular images acquired using Oppo.  The average difference in the accuracy between males and females is $14.68\%$ for right ocular images for Oppo. Lowest FPR ($15.49\%$) and FNR ($13.78\%$) are obtained for left ocular images under dark lighting conditions acquired using Oppo. The lowest FPR ($18.2\%$) and FNR ($6.8\%$) are obtained for left ocular images under office lighting conditions for Note-4. Our results are in contrary to those obtained in~\cite{9211002} where females outperformed males in gender classification based on ocular region. However, in this study~\cite{9211002} ocular regions are cropped from Labeled Faces in the Wild dataset.

\par
\begin{figure}[!h]
    \centering 
    \begin{subfigure}{0.2\textwidth}
      \includegraphics[width=\linewidth]{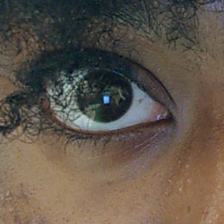}
      \caption{Obstruction}
      \label{fig:1}
    \end{subfigure}\hfil 
    \begin{subfigure}{0.2\textwidth}
      \includegraphics[width=\linewidth]{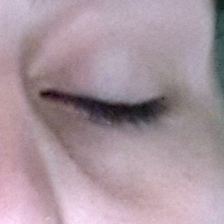}
      \caption{Closed eyelid}
      \label{fig:2}
    \end{subfigure}\hfil 
    \begin{subfigure}{0.2\textwidth}
      \includegraphics[width=\linewidth]{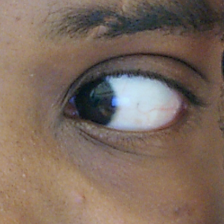}
      \caption{Gazing}
      \label{fig:3}
    \end{subfigure}
    
    \medskip
    \begin{subfigure}{0.2\textwidth}
      \includegraphics[width=\linewidth]{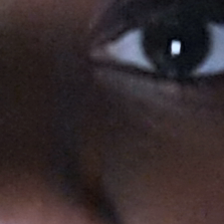}
      \caption{Poor lighting}
      \label{fig:4}
    \end{subfigure}\hfil 
    \begin{subfigure}{0.2\textwidth}
      \includegraphics[width=\linewidth]{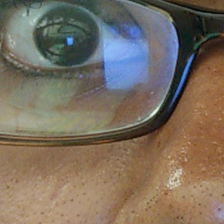}
      \caption{Eyeglasses}
      \label{fig:5}
    \end{subfigure}\hfil 
    \begin{subfigure}{0.2\textwidth}
      \includegraphics[width=\linewidth]{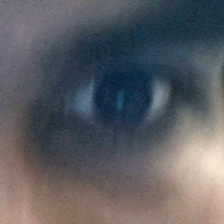}
      \caption{Motion blur}
      \label{fig:6}
    \end{subfigure}
    \caption{Example of covariates in ocular images of females, commonly available in mobile environment, and attributing to the error rate of the gender classifiers.}
    \label{fig:images}
    \end{figure}

 Further, based on manual inspection, we observed that \emph{covariates such as eye-gazing, eyeglasses, obstructions, the presence of hair, and low lighting to be the major factors contributing to the error rate of the gender classifier especially for females}. Figure \ref{fig:images} shows some of the female sample eye images misclassified by the gender classification models. 

    
\begin{table*}[]
\centering
    \caption{Gender classification accuracy rates of ResNet-$50$, MobileNet-V$2$ and their ensemble across males (M) and females (F) in different lighting conditions, when trained and test on left eye images acquired using Note-4.}
\begin{adjustbox}{width=1\textwidth}

\begin{tabular}{|l|c|c|c|c|c|c|c|c|c|}
\hline
\multicolumn{1}{|c|}{\textbf{}} & \multicolumn{3}{c|}{\textbf{ResNet-$50$}}              & \multicolumn{3}{c|}{\textbf{MobileNet-V$2$}}            & \multicolumn{3}{c|}{\textbf{Ensemble}}               \\ \hline
                                & M                & F              & Overall Acc.     & M                & F              & Overall Acc.      & M               & F               & Overall Acc.     \\ \hline
Dark                            & 98               & 59             & 78.71            & 83               & 79             & 80.88             & 93              & 74.5            & 83.93            \\ \hline
Daylight                        & 90.2             & 69             & 80.76            & 91.7             & 63             & 78.56             & 89.6            & 71.7            & 81.12            \\ \hline
Office                          & 94               & 64.6           & 79.69            & 93.5             & 76.6           & 85.43             & 94.2            & 78.9            & 87.01            \\ \hline
                                & \multicolumn{2}{c|}{\textbf{FPR}} & \textbf{FNR}     & \multicolumn{2}{c|}{\textbf{FPR}} & \textbf{FNR}      & \multicolumn{2}{c|}{\textbf{FPR}} & \textbf{FNR}     \\ \hline
Dark                            & \multicolumn{2}{c|}{29.4}         & 3.2              & \multicolumn{2}{c|}{20.3}         & 17.8              & \multicolumn{2}{c|}{21.4}         & 8                \\ \hline
Daylight                        & \multicolumn{2}{c|}{25.6}         & 12.4             & \multicolumn{2}{c|}{28.9}         & 11.7              & \multicolumn{2}{c|}{24}           & 12.6             \\ \hline
Office                          & \multicolumn{2}{c|}{27.3}         & 8.5              & \multicolumn{2}{c|}{20}           & 7.8               & \multicolumn{2}{c|}{18.2}         & 6.8              \\ \hline
\end{tabular}%
\end{adjustbox}
\label{left_Note4}
\end{table*}

\begin{table*}[]
\centering
    \caption{Gender classification accuracy rates of ResNet-$50$, MobileNet-V$2$ and their ensemble across males (M) and females (F) in different lighting conditions, when trained and tested on left eye images acquired using Oppo.}
\begin{adjustbox}{width=1\textwidth}

\begin{tabular}{|l|c|c|c|c|c|c|c|c|c|}
\hline
\multicolumn{1}{|c|}{\textbf{}} &
  \multicolumn{3}{c|}{\textbf{ResNet-$50$}} &
  \multicolumn{3}{c|}{\textbf{MobileNet-V$2$}} &
  \multicolumn{3}{c|}{\textbf{Ensemble}} \\ \hline
         & M            & F           & Overall Acc. & M            & F           & Overall Acc. & M            & F           & Overall Acc. \\ \hline
Dark     & 96.26        & 66.67       & 81.6         & 95.74        & 63.34       & 79.64        & 96.6         & 70          & 83.4         \\ \hline
Daylight & 91.76        & 60.75       & 77.31        & 93.2         & 58.05       & 76.71        & 92.07        & 62.45       & 77.41        \\ \hline
Office   & 95.6         & 61.07       & 78.61        & 97.25        & 56.18       & 77.11        & 97.23        & 62.94       & 80.47        \\ \hline
 &
  \multicolumn{2}{c|}{\textbf{FPR}} &
  \textbf{FNR} &
  \multicolumn{2}{c|}{\textbf{FPR}} &
  \textbf{FNR} &
  \multicolumn{2}{c|}{\textbf{FPR}} &
  \textbf{FNR} \\ \hline
Dark     & \multicolumn{2}{c|}{25.7}  & 5.3          & \multicolumn{2}{c|}{27.69} & 6.3          & \multicolumn{2}{c|}{23.72} & 4.64         \\ \hline
Daylight & \multicolumn{2}{c|}{29.96} & 11.9         & \multicolumn{2}{c|}{31.0 3} & 0.104        & \multicolumn{2}{c|}{28.96} & 11.26        \\ \hline
Office   & \multicolumn{2}{c|}{28.94} & 6.7          & \multicolumn{2}{c|}{31.05} & 4.65         & \multicolumn{2}{c|}{27.58} & 4.11         \\ \hline
\end{tabular}%
\end{adjustbox}
\label{left_Oppo}
\end{table*}	

\begin{table*}[]
\centering
   \caption{Gender classification accuracy rates of ResNet-$50$, MobileNet-V$2$ and their ensemble across males (M) and females (F) in different lighting conditions, when trained and tested on right eye images acquired using Note-4.}
\begin{adjustbox}{width=1\textwidth}
\begin{tabular}{|l|c|c|c|c|c|c|c|c|c|}
\hline
\multicolumn{1}{|c|}{\textbf{}} & \multicolumn{3}{c|}{\textbf{ResNet-50}}              & \multicolumn{3}{c|}{\textbf{MobileNet-V2}}            & \multicolumn{3}{c|}{\textbf{Ensemble}}               \\ \hline
                                & M                & F              & Overall Acc.     & M                & F              & Overall Acc.      & M               & F               & Overall Acc.     \\ \hline
Dark                            & 91.3             & 70.3           & 80.83            & 80.5             & 81.6           & 81.02             & 89.7            & 76.2            & 82.94            \\ \hline
Daylight                        & 95               & 57             & 77.59            & 92               & 74             & 84.03             & 95.5            & 61.7            & 79.32            \\ \hline
Office                          & 98.4             & 70.2           & 84.64            & 95.3             & 78.2           & 86.89             & 98.2            & 71              & 85.12            \\ \hline
                                & \multicolumn{2}{c|}{\textbf{FPR}} & \textbf{FNR}     & \multicolumn{2}{c|}{\textbf{FPR}} & \textbf{FNR}      & \multicolumn{2}{c|}{\textbf{FPR}} & \textbf{FNR}     \\ \hline
Dark                            & \multicolumn{2}{c|}{24.6}         & 11               & \multicolumn{2}{c|}{18.6}         & 19.3              & \multicolumn{2}{c|}{21}           & 12               \\ \hline
Daylight                        & \multicolumn{2}{c|}{31.3}         & 8.55             & \multicolumn{2}{c|}{21.9}         & 9.6               & \multicolumn{2}{c|}{28.6}         & 6.7              \\ \hline
Office                          & \multicolumn{2}{c|}{23.2}         & 2.2              & \multicolumn{2}{c|}{18.6}         & 5.7               & \multicolumn{2}{c|}{22.8}         & 2.5              \\ \hline
\end{tabular}%
\end{adjustbox}
\label{Right_Note4}
\end{table*}

\begin{table*}[]
\centering
    \caption{Gender classification accuracy rates of ResNet-$50$, MobileNet-V$2$ and their ensemble across males (M) and females (F) in different lighting conditions, when trained and tested on right eye images acquired using Oppo.}
\begin{adjustbox}{width=1\textwidth}

\begin{tabular}{|l|c|c|c|c|c|c|c|c|c|}
\hline
\multicolumn{1}{|c|}{\textbf{}} & \multicolumn{3}{c|}{\textbf{ResNet-50}}             & \multicolumn{3}{c|}{\textbf{MobileNet-V2}}           & \multicolumn{3}{c|}{\textbf{Ensemble}}              \\ \hline
                                & M               & F               & Overall Acc.    & M               & F               & Overall Acc.     & M                & F              & Overall Acc.    \\ \hline
Dark                            & 92.45           & 67.47           & 80.04           & 81.97           & 90.82           & 86.33            & 90.36            & 77.9           & 84.19           \\ \hline
Daylight                        & 91.26           & 66.3            & 79.22           & 85.03           & 81.44           & 83.7             & 89.94            & 72.96          & 81.49           \\ \hline
Office                          & 95.86           & 67.91           & 82.21           & 92.36           & 76.38           & 84.63            & 95.6             & 71             & 83.59           \\ \hline
                                & \multicolumn{2}{c|}{\textbf{FPR}} & \textbf{FNR}    & \multicolumn{2}{c|}{\textbf{FPR}} & \textbf{FNR}     & \multicolumn{2}{c|}{\textbf{FPR}} & \textbf{FNR}    \\ \hline
Dark                            & \multicolumn{2}{c|}{26.02}        & 10.05           & \multicolumn{2}{c|}{10.07}        & 16.57            & \multicolumn{2}{c|}{19.65}        & 11.01           \\ \hline
Daylight                        & \multicolumn{2}{c|}{26.97}        & 11.65           & \multicolumn{2}{c|}{17.91}        & 15.52            & \multicolumn{2}{c|}{23.11}        & 12.12           \\ \hline
Office                          & \multicolumn{2}{c|}{25.07}        & 5.74            & \multicolumn{2}{c|}{20.34}        & 9.086            & \multicolumn{2}{c|}{23.3}         & 5.85            \\ \hline
\end{tabular}%

\end{adjustbox}
\label{Right_Oppo}
\end{table*}

\section{Conclusion}
This paper evaluates the fairness of the mobile user authentication and gender classification algorithms based on ocular region across males and females. In contrary to the existing studies on face recognition, we obtained equivalent authentication performance for males and females based on the ocular region at lower FMR points $(1^{-4})$ and an overall Area Under Curve (AUC). The reason could be the robustness of the subject-specific templates of ocular region to facial expression change, make-up, and facial morphological differences over face biometrics. However, males outperformed females by a significant difference of $22.58\%$ in gender classification. This error rate was mainly due to the presence of covariates such as hair, eyeglasses, motion blur, and eye gazing. As a part of future work, experiments will be extended on other ocular biometric datasets captured in the near-infrared and visible spectrum across gender, race and age. The impact of the covariates and multi-frame fusion in unequal accuracy rates of the ocular-based gender classifiers will be quantified.

\section{Acknowledgment}
Rattani is the co-organizer of the IEEE ICIP 2016 VISOB 1.0 and IEEE WCCI 2020 VISOB 2.0 mobile ocular biometric competitions. Authors would like to thank Narsi Reddy and Mark Nguyen for their assistance in dataset processing.
%
%
\small{
 \bibliographystyle{splncs04}
 \bibliography{mybibliography,biblio,sample1,sample}
 }

\end{document}